\documentclass{article}
\usepackage{ijcai24}
\usepackage{pdfpages}
\usepackage{times}
\usepackage{soul}
\usepackage{url}
\usepackage[hidelinks]{hyperref}
\usepackage[utf8]{inputenc}
\usepackage[small]{caption}
\usepackage{graphicx}
\usepackage{amsmath}
\usepackage{amsthm}
\usepackage{booktabs}
\usepackage{algorithm}
\usepackage{algorithmic}
\usepackage[switch]{lineno}
\usepackage{multirow}
\usepackage{xcolor}

\usepackage{amssymb}
\usepackage{pifont}
\newcommand{\cmark}{\ding{51}}%
\newcommand{\xmark}{\ding{55}}%
\usepackage{threeparttable}

\title{Empowering Time Series Analysis with Large Language Models: A Survey}

\author{Yushan Jiang\textsuperscript{\rm 1,$\dag$}, Zijie Pan\textsuperscript{\rm 1,$\dag$}, Xikun Zhang\textsuperscript{\rm 2}, Sahil Garg\textsuperscript{\rm 3}, Anderson Schneider\textsuperscript{\rm 3}, \\
Yuriy Nevmyvaka\textsuperscript{\rm 3, *}, Dongjin Song\textsuperscript{\rm 1, *}
    \affiliations
    \textsuperscript{\rm 1}School of Computing, University of Connecticut, USA\\
    \textsuperscript{\rm 2}School of Computer Science, The University of Sydney, Australia\\
    \textsuperscript{\rm 3}Department of Machine Learning Research, Morgan Stanley, USA\\
    \emails{
    \{yushan.jiang, zijie.pan, dongjin.song\}@uconn.edu, 
    xzha0505@uni.sydney.edu.au,\\
    \{sahil.garg, anderson.schneider, yuriy.nevmyvaka\}@morganstanley.com,
    }
}

\begin{document}

\maketitle

\begin{NoHyper}
\def\thefootnote{}\footnotetext{$\dag$ equal contribution, * corresponding authors}
\end{NoHyper}

\begin{abstract}

Recently, remarkable progress has been made over large language models (LLMs), demonstrating their unprecedented capability in varieties of natural language tasks. 
However, completely training a large general-purpose model from the scratch is challenging for time series analysis, due to the large volumes and varieties of time series data, as well as the non-stationarity that leads to concept drift impeding continuous model adaptation and re-training. Recent advances have shown that pre-trained LLMs can be exploited to capture complex dependencies in time series data and facilitate various applications. In this survey, we provide a systematic overview of existing methods that leverage LLMs for time series analysis. Specifically, we first state the challenges and motivations of applying language models in the context of time series as well as brief preliminaries of LLMs. Next, we summarize the general pipeline for LLM-based time series analysis, categorize existing methods into different groups (\textit{i.e.}, direct query, tokenization, prompt design, fine-tune, and model integration), and highlight the key ideas within each group. We also discuss the applications of LLMs for both general and spatial-temporal time series data, tailored to specific domains. Finally, we thoroughly discuss future research opportunities to empower time series analysis with LLMs.

\end{abstract}

\section{Introduction}
In the past few years, significant advances have been made in large language models (LLMs), taking artificial intelligence and natural language processing a giant leap forward. LLMs, \textit{e.g.}, OpenAI's GPT-3 and Meta's Llama 2~\cite{touvron2023llama2}, have not only exhibited an unparalleled ability to create narratives that are both coherent and contextually relevant but also demonstrated their remarkable accuracy and proficiency in complex and nuanced tasks such as responding to queries, translating sentences between multiple languages, code generation, and so on.

\begin{figure}[t]
\begin{center}
\centerline{\includegraphics[width=0.99\columnwidth]{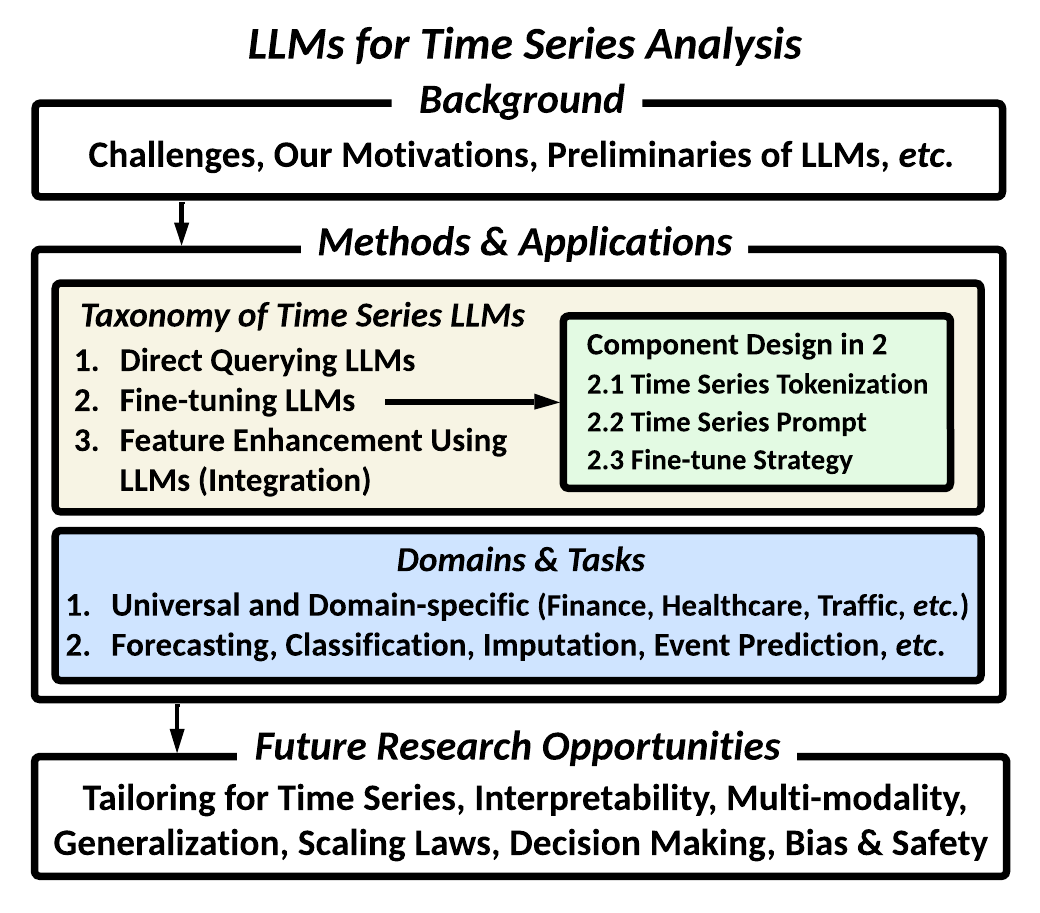}}
\vspace{-0.3cm}
\caption{The framework of our survey} 
\label{framework}
\end{center}
\vspace{-0.8cm}
\end{figure}

Inspired by the success of LLMs, a great deal of effort has been made to train general-purpose time series analysis models~\cite{wu2022timesnet,garza2023timegpt} to facilitate various underlying tasks, such as classification, forecasting, and anomaly detection. These efforts, however, are hindered by two key challenges. First, time series data may come in various forms - univariate or multivariate, for example - in large volumes, and from a variety of domains: healthcare, finance, traffic, environmental sciences, \textit{etc}. This increases the complexity of model training and makes it difficult to handle different scenarios. Second, real-world time series data often exhibit non-stationary properties when they are continuously accumulated/collected, meaning that the statistical characteristics of time series data, such as mean, variance, and auto-correlation, will change over time. This phenomenon is common in applications such as financial markets, climate data, and user behavior analytics where patterns and trajectories evolve and do not remain constant. It can lead to the concept drift problem, as the statistical properties of the target variables may also change over time, making it difficult for the large models to be continuously adapted and re-trained~\cite{kim2021reversible}. 

More recently, instead of training a general-purpose time series analysis model from the scratch, there is an increasing trend in exploiting existing LLMs in various time series applications. Consequently, different methodologies have been developed based on application types. In this survey, we provide a comprehensive and systematic overview of existing methods that leverage LLMs for time series analysis. As shown in Figure \ref{framework}, we will first discuss the challenges, motivations, and preliminaries of LLMs. Next, we will summarize the general pipeline for LLM-based time series analysis and introduce five different types of techniques for applying LLMs: direct query, tokenization, prompt design, fine-tuning, and model integration. We will also discuss application of LLMs to specific domains. For better comparison, we provide a comprehensive table that summarizes representative methods, their modeling strategies, associated tasks and domains (as shown in Table~\ref{tab:taxonomy}). Finally, we highlight potential future research opportunities to further advance time series analysis with LLMs. In summary, the main contributions of this survey include:
\begin{itemize}
    \item We catalog papers on LLM-based time series analysis that cover 21 representative approaches since 2022.
    \item We systematically survey existing methods that leverage LLMs for time series analysis, uniquely categorize them into five groups based on the methodology, and discuss their application tasks and domains.
    \item We discuss and highlight future directions that advance time series analysis with LLMs and encourage researchers and practitioner to further investigate this field.
\end{itemize}

\section{Background}

\subsection{General Large Language Models}

Early advancements in natural language processing involve neural language models (NLMs)~\cite{arisoy2012deep} and pioneering LLMs, such as GPT-2~\cite{radford2019language}, BERT~\cite{devlin2018bert}, RoBERTa~\cite{liu2019roberta}, and XLNet~\cite{yang2019xlnet}. More recently, the rise of more powerful LLMs (\textit{e.g.,} multi-modal large language models~\cite{yin2023survey}) has revolutionized AI usage because of their exceptional ability to handle complex tasks. We adopt a similar criterion to that in~\cite{zhao2023survey,jin2023large} and divide LLMs into two categories: embedding-visible LLMs and embedding-invisible LLMs. The embedding-visible LLMs are usually open-sourced with inner states accessible. Notable examples include T5~\cite{raffel2020exploring}, Flan-T5~\cite{chung2022scaling}, LLaMA~\cite{touvron2023llama1,touvron2023llama2}, ChatGLM~\cite{du2022glm}, \textit{etc}. These open-sourced LLMs are adaptable for various downstream tasks, demonstrating impressive capabilities in both few-shot and zero-shot learning settings, without the need to retrain from scratch. On the other hand, the embedding-invisible LLMs are typically closed-sourced with inner
states inaccessible to the public. This type of LLMs include PaLM~\cite{chowdhery2023palm}, GPT-3~\cite{brown2020language}, GPT-4~\cite{Achiam2023GPT4TR}. For these models, researchers are limited to conducting inference tasks through prompting via the API calls. 
These LLMs can be potentially exploited for time series analysis. 

\subsection{Leveraging LLMs in Time Series Analysis}

The rapid development of LLMs in natural language processing has unveiled unprecedented capabilities in sequential modeling and pattern recognition. It is natural to ask: How can LLMs be effectively leveraged to advance general-purpose time series analysis? 

Our survey aims to answer the question based on a thorough overview of existing literature. We claim that LLMs can serve as a flexible as well as highly competent component in the time series modeling. The flexibility lies in a wide spectrum of available LLMs that can be employed and the variety of ways they can be configured for time series analysis (Section 3). Regarding their competence, LLMs can be tailored for a wide range of real-world applications with domain-specific context (Section 4). Certainly, there still exists several challenges in this field and we discuss future opportunities (Section 5).

Next, we highlight the difference between our survey and a few recent relevant ones in terms of the scope and focus. Deldari~[\citeyear{deldari2022beyond}] and Ma \textit{et al.}~[\citeyear{ma2023survey}] both include the summary of pre-trained techniques for time series where Deldari~[\citeyear{deldari2022beyond}] is specialized in self-supervised representation learning (SSRL) methods for multi-modal temporal data (not only time series). Mai \textit{et al.}~[\citeyear{mai2023opportunities}] summarizes large pre-trained models (including LLMs) for time series in geospatial domain. Jin \textit{et al.}~[\citeyear{jin2023large}] provides a comprehensive survey of large pre-trained models for time series and general spatial-temporal data. Compared with \cite{jin2023large,mai2023opportunities}, our survey focuses on \textbf{LLMs} for \textbf{time series analysis}, which is the only one categorizing existing methods based on modeling strategy. Our survey is also uniquely positioned to provide detailed introductions of not only universal methodology design but also various applications with domain-specific context. Figure~\ref{categorization} and Table~\ref{tab:taxonomy} demonstrate our uniqueness.

\section{Taxonomy of LLMs in Time Series Analysis}

In this section, we conduct a detailed discussion of existing research that utilizes LLMs for universal time series modeling and thoroughly analyze the design of their components, where we categorize and brief the designs of domain-specific methods. We will also elaborate by tailoring them to specific domain contexts in Section 4. The detailed taxonomy is provided in Table~\ref{tab:taxonomy}.

\begin{figure}[ht]
\begin{center}
\centerline{\includegraphics[width=0.98\columnwidth]{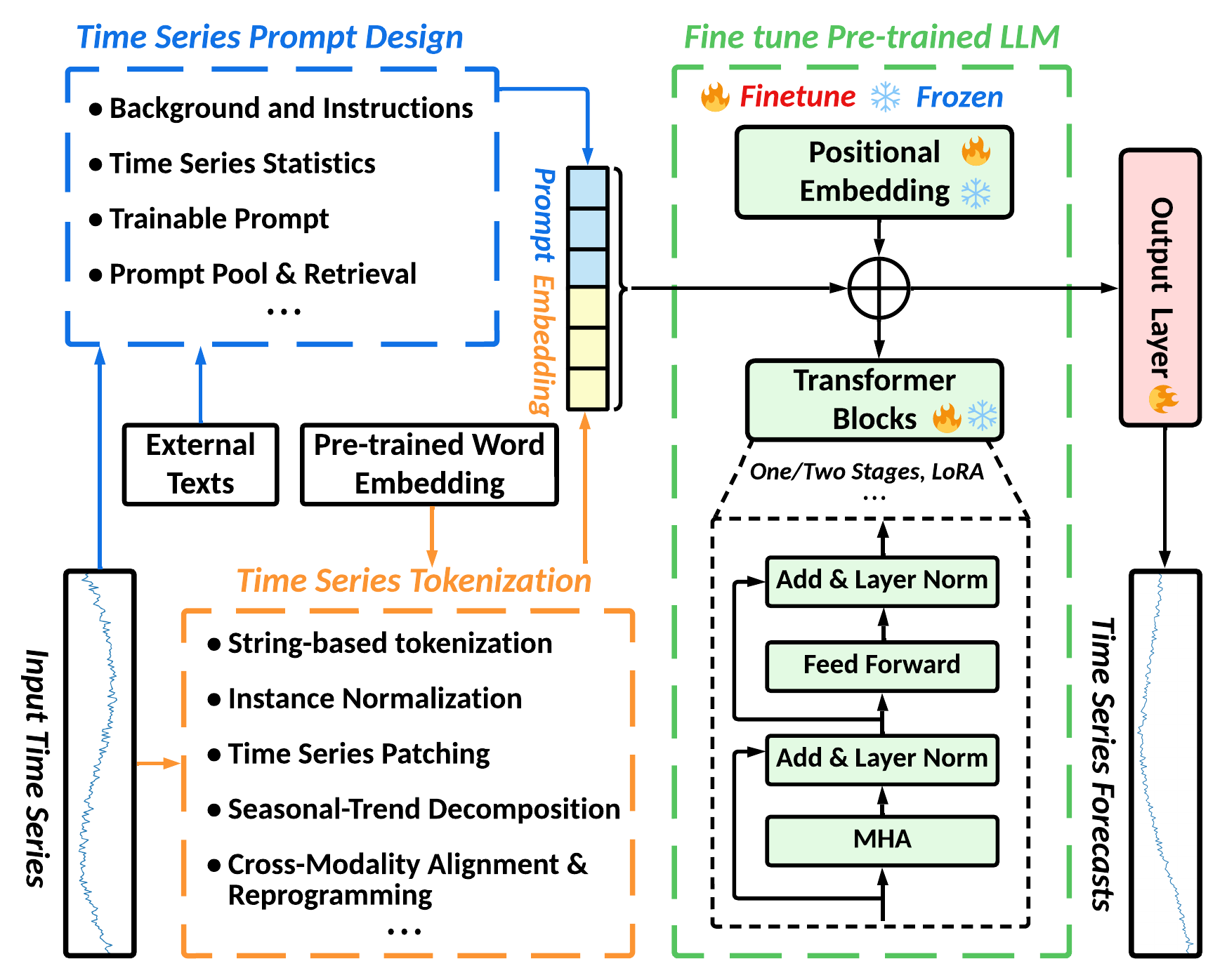}}
\vspace{-0.2cm}
\caption{Categorization of component design for fine-tuning time series LLMs (Section 3.2-3.4)} 
\label{categorization}
\end{center}
\vspace{-0.5cm}
\end{figure}

\medskip

\noindent \textbf{General Pipeline of LLMs}. To adopt LLMs for time series analysis, three primary methods are employed: direct querying of LLMs (Section 3.1), fine-tuning LLMs with tailored designs (Section 3.2-3.4), and incorporating LLMs into time series models as a means of feature enhancement (Section 3.5). Specifically, three key components can be leveraged to fine-tune LLMs as shown in Figure~\ref{categorization}: The input time series are first tokenized into embedding based on proper tokenization techniques, where proper prompts can be adopted to further enhance the time series representation. As such, LLMs can better comprehend prompt-enhanced time series embedding and be fine-tuned for downstream tasks, based on sophisticated strategies.

\subsection{Direct Query of LLMs}

PromptCast~\cite{xue2023promptcast} is the first work that directly conducts general time series forecasting in sentence-to-sentence fashion using pre-trained LLMs. It introduces a novel forecasting setting, \textit{i.e.}, prompt-based time series forecasting that embeds lag information as well as instructions into the prompts and uses output sentences from the LLMs to conduct forecasting. Directly querying LLMs can also be beneficial in domain-specific scenarios ~\cite{yu2023temporal,wang2023i}, particularly when leveraging advanced pre-trained LLMs (\textit{e.g.,} GPT-4~\cite{Achiam2023GPT4TR} and OpenLLaMA ~\cite{openlm2023openllama,together2023redpajama,touvron2023llama2}) in conjunction with context-inclusive prompts that provide relevant domain knowledge. 

While direct usage of LLMs for time series forecasting can be zero-shot or few-shot, instruction-based fine-tuning, and Chain-of-Thoughts (COT)~\cite{lightman2023let,wei2022chain,zhang2023multimodal} have shown positive effects on the reasoning process. LLMTime~\cite{gruver2023llmtime} also demonstrates that LLMs are effective zero-shot time series learners with proper text-wise tokenization on time series.

\subsection{Time Series Tokenization Design}

The aforementioned works~\cite{xue2023promptcast,yu2023temporal,gruver2023llmtime} convert numerical values of time series data into string-based tokens so that LLMs can seamlessly encode time series as the natural language inputs. In this subsequent section, we exclusively focus on the tokenization design to represent time series data more effectively. In practical applications, time series analysis often encounters the challenge of distribution shifts. To address this issue, major works adopt channel independence and reversible instance normalization (RevIN)~\cite{kim2021reversible} before tokenizing the time series data.

Patch representation~\cite{Yuqietal-2023-PatchTST} for time series has shown promising results in time series analysis with transformer-based models. For a univariate time series input with length L: $\mathbf{X}_{1 \mathrm{D}} \in \mathbb{R}^L$, the patching operation first repeats the final value in the original univariate time series S times. Then, it unfolds the input univariate time series through a sliding window with the length of patch size $P$ and the stride size of $S$. Through patching, the univariate time series will be transformed into two-dimensional representations $\mathbf{X}_{\text{p}}\in \mathbb{R}^{P \times N}$, where $N$ is the number of patches with $N=\left\lfloor\frac{(L-P)}{S}\right\rfloor+2$. Patching can be mathematically formulated as:
\begin{equation}
{\mathbf{X}}_{\text{p}}=\text{Unfold}\left(\text{Right\_Pad}\left({\mathbf{X}}_{1 \mathrm{D}}\right),\text{size}=P,\text{stride}=S\right)
\end{equation}
Patching tokenization design preserves the original relative order of the data and aggregates local information into each patch. One Fits All (OFA)~\cite{zhou2023onefitsall}, LLM4TS~\cite{chang2023llm4ts}, TEST~\cite{sun2023test}, TEMPO~\cite{cao2023tempo}, and Time-LLM~\cite{jin2023time} primarily adopt this method to tokenize time series data. In order to harmonize the modalities of numerical data and natural language, an auxiliary loss has been introduced by TEST~\cite{sun2023test} to enhance the cosine similarity between the embeddings of time series tokens and selected text prototypes, as well as to ensure proximity in the textual prototype space for similar time series instances. For a similar purpose, Time-LLM~\cite{jin2023time} proposes to use a multi-headed attention mechanism to align the patched time series representation with the pre-trained text prototype embedding, acquired through linear probing. Specifically, Time-LLM reprograms time series patches in each attention head via:
\begin{align}
\mathbf{Z}_k^{(i)} &= \operatorname{Attention}(\mathbf{Q}_k^{(i)}, \mathbf{K}_k^{(i)}, \mathbf{V}_k^{(i)}) \nonumber \\
                   &= \operatorname{Softmax}\left(\frac{\mathbf{Q}_k^{(i)} \mathbf{K}_k^{(i) \top}}{\sqrt{d_k}}\right) \mathbf{V}_k^{(i)}
\end{align}
where query matrices $\mathbf{Q}_k^{(i)}=\hat{\mathbf{X}}_p^{(i)} \mathbf{W}_k^Q$, key matrices $\mathbf{K}_k^{(i)}=\mathbf{E}^{\prime} \mathbf{W}_k^K$, value matrices $\mathbf{V}_k^{(i)}=\mathbf{E}^{\prime} \mathbf{W}_k^V$, and $\mathbf{E}^{\prime}$ is the reduce-sized pre-trained word embedding.

Moreover, Prompt-based Generative Pre-trained Transformer (TEMPO)~\cite{cao2023tempo} uses the additive STL decomposition~\cite{cleveland1990stl} to extract trend, seasonal, and residual components to better tokenize input time series, \textit{i.e.}, $\mathbf{X}_{\text{p}} = \mathbf{X}_{\text{p}}^{T} + \mathbf{X}_{\text{p}}^{S} + \mathbf{X}_{\text{p}}^{R}$, where the trend term $\mathbf{X}_{\text{p}}^{T}$ represents the long-term pattern, the seasonal component $\mathbf{X}_{\text{p}}^{S}$ contains the recurring short-term seasonal patterns and $\mathbf{X}_{\text{p}}^{R}$ includes the residuals of the time series data after removing trend and seasonal components.

\subsection{Prompt Design}
PromptCast~\cite{xue2023promptcast} develops template-based prompts for LLM time series forecasting, while some methods~\cite{yu2023temporal,xue2022leveraging,wang2023i,liu2023large} enrich the prompt design by incorporating LLM-generated or gathered background information, which highlights the importance of context-inclusive prompts in real-world applications. Besides the background and instruction prompts, Time-LLM~\cite{jin2023time} adds statistical information of the time series data to facilitate time series forecasting. 
Compared with fixed and non-trainable prompts, a soft and trainable prompt makes it easier for LLMs to understand and align with the input~\cite{lester2021power}. Prefix soft prompts are the task-specific embedding vectors, learned based on a loss from LLMs' output and the ground truth.

TEST~\cite{sun2023test} initializes the soft prompts with uniform distributions, text embedding of the downstream task labels, or the most common words from
the vocabulary. TEMPO~\cite{cao2023tempo} focuses on retrieval-based prompt design. Similar to L2P~\cite{wang2022learning}, it first introduces a shared pool of prompts stored as distinct key-value pairs and then selects the most representative soft prompts for fine-tuning via a similarity score matching mechanism.        

\subsection{Fine-tuning Strategy}

Fine-tuning pre-trained LLMs is pivotal to leveraging LLMs' strong pattern recognition and reasoning capabilities to facilitate downstream tasks. Several existing works opt to fine-tune pre-trained LLMs directly (one stage) for time series analysis. The main difference lies in how the modules' parameters are updated during the fine-tuning process. As a standard practice~\cite{lu2022frozen,houlsby2019parameter}, OFA~\cite{zhou2023onefitsall} fine-tunes the positional embedding and layer normalization layers, and freezes self-attention layers and Feedforward Neural Networks (FFN) as they contain majorities of the learned knowledge. LLM4TS~\cite{chang2023llm4ts} and TEMPO ~\cite{cao2023tempo} further fine-tune the self-attention modules using Low-Rank Adaptation (LoRA)~\cite{hu2021lora} by introducing trainable low-rank bypasses to the query ($Q$) and key ($K$) matrices in the self-attention mechanism. Instead of directly modifying the original weight matrices $W^{Q}$ and $W^{K}$, LoRA introduces $A^{Q}$, $A^{K}$, $B^{Q}$, and $B^{K}$, which are much smaller in size compared to $W^{Q}$ and $W^{K}$. The modified query and key matrices in LoRA can be represented as:
\begin{equation}
\begin{aligned}
& \operatorname{LoRA}(Q)=X W^Q+X B^Q A^Q \\
& \operatorname{LoRA}(K)=X W^K+X B^K A^K
\end{aligned}
\end{equation}
where $A^{Q}$, $A^{K}$ are the trainable low-rank matrices, and $B^{Q}$, and $B^{K}$ are the projection matrices that project the input $X$ into a lower-dimensional space. The addition of these low-rank matrices to the original query and key matrices allows the model to fine-tune more effectively and efficiently with fewer trainable parameters. Besides one-stage fine-tuning, LLM4TS~\cite{chang2023llm4ts} proposes a two-stage fine-tuning strategy to accommodate LLMs to time series data. The first stage is supervised autoregressive fine-tuning, in which the backbone model predicts contiguous patches based on a sequence of patches as the input. Moreover, during the second stage half epochs are set to train the final linear layer, and half epochs are set to train all parameters jointly, tailored to specific downstream forecasting tasks. 

\subsection{Integrating LLMs in Time Series Models}
Rather than directly querying or fine-tuning time series LLMs to generate output, some studies use frozen LLMs as a component that is inherently capable of enhancing the feature space of time series.
A frozen LLM can serve as a highly capable function in multi-stage modeling that provides intermediate processing of data or the output of the preceding component, and feeds it to the subsequent neural networks~\cite{xue2022leveraging,shi2023language} or regression analysis~\cite{lopez2023can}.
Specifically, LLMs can be efficiently applied within a multimodal self-supervised framework for time series analysis. Here, embeddings from time series data and LLM-generated text embeddings are used as positive and negative pairs to refine the model through contrastive loss optimization~\cite{sun2023test,li2023frozen}.
Because of LLMs' inherent capability to understand natural language, they are also a good fit for generating complex inter-series dependencies for downstream multivariate time series modeling whenever external related text is available. LA-GCN~\cite{xu2023language} and Chen \textit{et al.}~[\citeyear{chen2023chatgpt}] use LLMs to learn the topological structure of multivariate time series from domain-specific text.

\section{Applications of Time Series LLMs}
In this section, we review the existing applications of LLMs to general and spatial-temporal time series data, which covers universal and domain-specific areas including finance, transportation, healthcare, and computer vision. 
 
\begin{table*}[t] 
  \setlength{\tabcolsep}{4.3pt}
  \renewcommand{\arraystretch}{0.75}
  \scriptsize
  \caption{\footnotesize{Taxonomy of time series LLMs. The data type \textbf{TS} denotes general time series, \textbf{ST} denotes spatial-temporal time series, the prefix \textbf{M-} indicates multi-modal inputs. The task entry \textbf{Multiple} includes forecasting, classification, imputation and anomaly detection. \textbf{Query} denotes direct query the whole LLMs for output, \textbf{Token} denotes the design of time series tokenization, \textbf{Prompt} indicates the design of textual or parameterized time series prompts, \textbf{Fine-tune} indicates if the parameters of LLMs are updated, \textbf{Integrate} indicates if LLMs are integrated as part of final model for downstream tasks. Code availability is assessed on January 31st, 2024.}}
  
  \vspace{-0.25cm}
  \centering
  \begin{threeparttable}
 
     \begin{tabular}{lcccccccccl}
     \toprule
     \cmidrule{1-11}
     
    \multirow{2}{*}{Method} & \multirow{2}{*}{Data Type} & \multirow{2}{*}{Domain} & \multirow{2}{*}{Task} &\multicolumn{5}{c}{Modeling Strategy} & \multirow{2}{*}{LLM} & \multirow{2}{*}{Code}  \\
    \cmidrule(lr){5-9}
    
     &   &  & & Query & Token & Prompt & Fine-tune & Integrate &   \\
    \midrule
    \cmidrule{1-11}
      
     Time-LLM~\cite{jin2023time} & M-TS  & General & Forecasting & \xmark & \cmark & \cmark & \cmark & \xmark & LLaMA, GPT-2 & Yes$^{\text{[1]}}$\\ 
     \midrule
     OFA~\cite{zhou2023onefitsall} &  TS & General & Multiple  & \xmark  & \cmark & \xmark & \cmark & \xmark & GPT-2 & Yes$^{\text{[2]}}$ \\ 
 
     \midrule
     TEMPO~\cite{cao2023tempo} & TS & General & Forecasting & \xmark & \cmark & \cmark & \cmark & \xmark &  GPT-2 & No \\ 

     \midrule
     \multirow{2}{*}{TEST~\cite{sun2023test}} & \multirow{2}{*}{M-TS} &  \multirow{2}{*}{General} & Forecasting & \multirow{2}{*}{\xmark} & \multirow{2}{*}{\cmark} & \multirow{2}{*}{\cmark} & \multirow{2}{*}{\xmark} & \multirow{2}{*}{\cmark} &  BERT, GPT-2  &  \multirow{2}{*}{Yes$^{\text{[3]}}$} \\
      &   &   &  Classification & & & & & & {{ChatGLM}}, LLaMA2 \\

     \midrule
     LLM4TS~\cite{chang2023llm4ts} & TS & General & Forecasting & \xmark & \cmark & \xmark & \cmark & \xmark &  GPT-2 & No\\ 

     \midrule
     PromptCast~\cite{xue2023promptcast} & TS & General & Forecasting & \cmark  & \xmark  & \cmark  & \xmark & \xmark & {Bart, BERT, \textit{etc.}} &Yes$^{\text{[4]}}$ \\ 

     \midrule
     LLMTIME~\cite{gruver2023llmtime} & TS & General & Forecasting & \cmark  & \cmark  & \xmark  & \xmark & \xmark & {GPT-3, LLaMA-2} & Yes$^{\text{[5]}}$  \\ 

     \midrule

     LAMP~\cite{shi2023language} & TS & General & Event Prediction & \cmark & \xmark  & \cmark & \xmark & \cmark & {GPT-3\&3.5, LLaMA-2} & Yes$^{\text{[6]}}$ \\

     \midrule
     \cite{gunjal2023drafting} & TS & General & Event Prediction & \cmark & \xmark & \cmark & \xmark & \xmark & {GPT-3.5, Flan-T5, \textit{etc.}} & No \\
     
     \midrule
     \cite{yu2023temporal} &  M-TS & Finance & Forecasting & \cmark &\xmark  & \cmark & \cmark & \xmark & {GPT-4, Open-LLaMA} &No  \\
     
      \midrule
     \cite{lopez2023can} & M-TS & Finance & Forecasting & \cmark & \xmark & \cmark & \xmark & \cmark & ChatGPT & No \\ 
     
     \midrule
     \cite{xie2023wall} & M-TS & Finance & Classification & \cmark & \xmark & \cmark & \xmark & \xmark & ChatGPT & No  \\
     
     \midrule
     \cite{chen2023chatgpt}  & M-TS & Finance & Classification & \xmark & \xmark & \cmark & \xmark  & \cmark & ChatGPT & Partial$^{\text{[7]}}$  \\ 
     
    \midrule
     METS~\cite{li2023frozen} & M-TS & Healthcare & Classification & \cmark  & \xmark  & \cmark  & \xmark & \cmark & ClinicalBERT & No   \\

     \midrule
    \cite{jiang2023health} & M-TS & Healthcare & Classification & \xmark  & \xmark & \xmark & \cmark & \xmark & NYUTron(BERT) & Yes$^{\text{[8]}}$  \\ 
     
     \midrule

    \multirow{2}{*}{\cite{liu2023large}} & \multirow{2}{*}{M-TS} &  \multirow{2}{*}{Healthcare} & Forecasting & \multirow{2}{*}{\cmark} & \multirow{2}{*}{\xmark} & \multirow{2}{*}{\cmark} & \multirow{2}{*}{\cmark} & \multirow{2}{*}{\xmark} & \multirow{2}{*}{PaLM}& \multirow{2}{*}{No} \\
      &   &   &  Classification  \\

     \midrule
     \multirow{2}{*}{AuxMobLCast~\cite{xue2022leveraging}} & \multirow{2}{*}{ST} & \multirow{2}{*}{Mobility} & \multirow{2}{*}{Forecasting} & \multirow{2}{*}{\xmark}  & \multirow{2}{*}{\xmark}  & \multirow{2}{*}{\cmark}  & \multirow{2}{*}{\cmark}  & \multirow{2}{*}{\cmark} & {BERT, RoBERTa} & Yes$^{\text{[9]}}$\\
     &&&&&&&&&{GPT-2, XLNet}\\
     
     \midrule
     LLM-Mob~\cite{wang2023i} & ST & Mobility & Forecasting & \cmark & \xmark & \cmark & \xmark & \xmark & GPT-3.5 & Yes$^{\text{[10]}}$   \\
     
     \midrule
     ST-LLM~\cite{liu2024spatialtemporal} & ST & Traffic & Forecasting &\xmark  & \cmark & \xmark & \cmark & \xmark & LLaMA, GPT-2 & No  \\

     \midrule
     GATGPT~\cite{chen2023gatgpt} & ST & Traffic & Imputation & \xmark & \cmark & \xmark & \cmark & \xmark & GPT-2 & No \\
     \midrule
     LA-GCN~\cite{xu2023language} & M-ST & Vision & Classification & \xmark & \cmark & \xmark & \xmark & \cmark & BERT & Yes$^{\text{[11]}}$ \\
    \bottomrule
    \cmidrule{1-11}
    \end{tabular}
    \begin{tablenotes}
        \tiny
         \item{} [1] https://github.com/kimmeen/time-llm \quad [2] https://github.com/DAMO-DI-ML/NeurIPS2023-One-Fits-All \quad %
         [3] https://openreview.net/forum?id=Tuh4nZVb0g (supplementary material) 
         
         \item{} [4] https://github.com/HaoUNSW/PISA \quad [5] https://github.com/ngruver/llmtime \quad [6] https://github.com/iLampard/lamp \quad[7] https://github.com/ZihanChen1995/ChatGPT-GNN-StockPredict 
         
         \item{} [8] https://github.com/nyuolab/NYUTron \quad [9] https://github.com/cruiseresearchgroup/AuxMobLCast \quad [10] https://github.com/xlwang233/LLM-Mob \quad [11] https://github.com/damNull/LAGCN
      \end{tablenotes}

    \end{threeparttable}
 \vspace{-0.4cm}
 \label{tab:taxonomy}
\end{table*}

\subsection{General Time Series Analysis}

\subsubsection{Universal Applications}
The aforementioned time series LLMs have been evaluated on a wide spectrum of benchmark datasets covering energy, traffic, electricity, weather, illness, business, aeronautics, and security~\cite{zhou2023onefitsall,sun2023test,gruver2023llmtime,xue2023promptcast,cao2023tempo,chang2023llm4ts,spathis2023first,jin2023time}. The tasks include forecasting, classification, imputation, and anomaly detection. These universal modeling methods can be tailored to each of these domains with specific knowledge.

While these applications are designed for structured time series data, a few recent studies have explored LLMs for a type of naturally observed temporal data with irregularities - the event sequence data. LAMP~\cite{shi2023language} first proposes to integrate an event prediction model with an LLM that performs abductive reasoning on real-world events. In the proposed framework, event candidate predictions are generated from historical event data (time, subject, and object) using a pre-trained base event sequence model, and an LLM is prompted to suggest possible cause events. This step is instruction-tuned with a few expert-annotated examples. For retrieval of relevant events, these events will be constructed as embeddings and matched against past events based on cosine similarity scores. Finally, an energy function with a continuous-time Transformer~\cite{xue2022hypro} learns to rank predictions with scores and output the event with the strongest retrieved evidence. The proposed framework outperforms state-of-the-art event sequence models on real-world benchmarks, indicating the superior performance of event reasoning via LLMs. Similarly, Gunjal and Durrett~[\citeyear{gunjal2023drafting}] attempts to use an LLM to construct structured representations of event knowledge (schema) directly in natural language, to achieve high recall over a set of human-curated events. In the experiments, multiple LLMs are considered and schemas are evaluated on different datasets, which highlights the importance of designing complex prompts for higher event coverage.

\subsubsection{Finance}
A recent trend in existing literature highlights the emergence of LLMs specialized for financial applications. Yu \textit{et al.}~[\citeyear{yu2023temporal}] focuses on a stock return prediction task by incorporating multi-modal data including the historical stock price, generated company profiles, and summarized weekly top news from GPT-4. Based on the designed prompt, this paper tests both instruction-based zero-shot/few-shot queries (with an effective alternative using COT approach) on GPT-4 and instruction-based fine-tuning on Open LLaMA. Results demonstrate that fine-tuned LLMs are capable of making decisions by analyzing information across modalities of financial data, thereby extracting meaningful insights and yielding explainable forecasts.
Similarly, Lopez-Lira and Tang~[\citeyear{lopez2023can}] directly queries ChatGPT and other large language models for stock market return predictions by using news headlines. A linear regression of the next day’s stock return is conducted on the recommendation score. A positive correlation between the scores and subsequent returns is observed, demonstrating the potential of LLMs to comprehend and forecast financial time series. 

Xie \textit{et al.}~[\citeyear{xie2023wall}] conducts an extensive study that queries ChatGPT (with designed prompts and COT alternatives) to test its zero-shot capabilities for multi-modal stock movement prediction. The experiments are conducted on three benchmark datasets that contain both stock prices and tweet data. Interestingly, even if ChatGPT demonstrates its effectiveness, the performance varies across datasets and underperforms even simple traditional methods. Observed limitations suggest the need for specialized fine-tuning techniques (\textit{e.g.,} the aforementioned Yu \textit{et al.}~[\citeyear{yu2023temporal}]) in the financial context.

Beyond direct queries or fine-tuning, Chen \textit{et al.}~[\citeyear{chen2023chatgpt}] proposes a framework that uses an external LLMs as a feature enhancement module for multi-modal stock movement prediction. Specifically, ChatGPT first generates an evolving graph structure of companies via prompting of news headlines at each time step, after which the static features of companies, the inferred structure, and historical stock prices are fed into a GNN and an LSTM for price movement prediction. This paper also provides an evaluation of portfolio performance with higher annualized cumulative returns, lower volatility, and lower maximum drawdown, suggesting the efficacy of LLMs in financial applications. 
 
We also acknowledge recent research efforts to develop financial LLMs where text input includes temporal information~\cite{wu2023bloomberggpt,xue2023weaverbird,zhang2023instruct}. These models, however, are more NLP-centric, with tasks including financial sentiment analysis, financial Q\&A, and named entity recognition. These studies are thus less relevant to our survey.  

\subsubsection{Healthcare}
Recent studies in healthcare have highlighted LLMs' capability to comprehend multi-modal medical context including physiological and behavioral time series, such as EEG (Electroencephalogram), ECG (Electrocardiogram), and Electronic Health Records (EHRs). METS~\cite{li2023frozen} framework aims to integrate LLMs into an ECG encoder classification. The model contains an ECG encoder based on ResNet1d-18, and a frozen large clinical language model, ClinicalBert~\cite{huang2019clinicalbert} that is pre-trained on all text from the MIMIC III dataset. A multi-modal self-supervised learning framework is used to align the paired ECG and text reports from the same patient while contrasting the unpaired ones via cosine similarity. In the zero-shot testing stage, the medical diagnostic statements constructed from discrete ECG labels are fed to ClinicalBert, and the similarity between ECG embedding and text embedding will be used for ECG classification. This paper first demonstrates the effectiveness of LLM-based self-supervised learning in multi-modality medical contexts.

Jiang \textit{et al.}~[\citeyear{jiang2023health}] proposes to develop an all-purpose clinical LLM, \textit{i.e.}, NYUTron, which is trained on EHRs and subsequently fine-tuned on three common clinical tasks and two operational tasks, such as the prediction of readmission, in-hospital mortality, comorbidity index, length of stay, as well as the insurance denial status. In this framework, clinical notes and task-specific labels are queried from the NYU Langone EHR database, which are used to pre-train a BERT model with masked language modeling objectives and perform subsequent fine-tuning. The trained model demonstrates improvements compared to traditional benchmarks on all tasks, suggesting the generalization capability of LLMs trained on clinical text. Note that a similar work~\cite{yang2022large} also aims to build a large clinical language model from scratch, but is more tailored to clinical NLP tasks, thus is not central to our discussion.
 
In addition to the developments of healthcare LLMs, healthcare datasets have given us other insights. For example, Liu \textit{et al.}~[\citeyear{liu2023large}] tests a pre-trained PaLM~\cite{chowdhery2023palm} on wearable and medical sensor recordings with three settings, zero-shot, prompt engineering, and prompt tuning for multiple healthcare tasks (cardiac signal analysis, physical activity recognition, metabolic calculation, and mental health). Their results emphasize the importance of healthcare time series for improving the capability of medical language models in the few-shot setting. Similarly, Spathis and Kawsar~[\citeyear{spathis2023first}] provide a case study of the tokenization of popular LLMs on mobile health sensing data, where a modality gap is identified and potential solutions such as prompt tuning, model grafting that maps time series via trained encoders onto the same token embedding space as text, as well as the design of new tokenizers for mixed time series and textual data are developed.

\subsection{Spatial-Temporal Time Series Analysis}

\subsubsection{Traffic}
In ST-LLM~\cite{liu2024spatialtemporal}, a spatial-temporal tokenization component is proposed so that an LLM is tailored for traffic forecasting tasks, where the input with exogenous information (\textit{e.g.,} hour of day, day of week) is encoded and integrated through point-wise convolutions and linear projections. Furthermore, the partial frozen training strategy is leveraged, with the multi-head attention in the last a few layers unfrozen in the fine-tune process to effectively handle spatial-temporal dependencies. Besides the general setting, the ST-LLM demonstrates advantages in terms of few-shot and zero-shot forecasting scenarios.
In addition to forecasting, the spatial-temporal imputation is initially explored by
GATGPT~\cite{chen2023gatgpt} that leverages a given topology of traffic networks together with LLMs. It exploits a trainable graph attention module~\cite{velivckovic2018graph} to enhance the embedding of irregular spatial-temporal inputs, which is processed by 1D convolution with additional positional encoding. As such, the pre-trained LLM can comprehend a given spatial correlation and be fine-tuned for imputation tasks.

\subsubsection{Human Mobility} 
Xue \textit{et al.}~[\citeyear{xue2022leveraging}] first leverages a non-numerical paradigm to perform spatial-temporal forecasting on human mobility data. Specifically, a mobility prompt, consisting of contextual Place-of-Interest (POI), temporal information, and mobility data, is designed and used to query the pre-trained LLM encoder, based on which the prompt embedding and the numerical token of mobility data is used to fine-tune the decoder to generate the token of prediction. An auxiliary POI category classification task built on top of a fully connected layer helps regularize the model training toward contextual forecasting and improve performance.
Instead of using prompt embedding as feature enhancement, LLM-Mob~\cite{wang2023i} directly queries an LLM for not only human movement prediction but also explanations based on an elaborated prompt. It integrates domain-specific knowledge of both long-term (\textit{i.e.}, historical stays) and short-term mobility patterns (\textit{i.e.}, the most recent movements) into the design of context-inclusive prompts. LLMs are guided to comprehend the underlying context of mobility data, and generate accurate forecasts as well as reasonable explanations without baseless speculations.

\subsubsection{Computer Vision}
One recent study of skeleton-based action recognition~\cite{xu2023language} also exhibits the importance of LLMs as an effective feature enhancement method in computer vision. Motivated by the potential of LLMs to capture the underlying knowledge, provide reasoning, and analyze actions within a skeleton sequence, this paper integrates LLM to generate faithful structural priors and action relations to assist spatial-temporal modeling. The names of all joints and those of all action labels are fed into a pre-trained BERT to get the text embeddings, based on which, each edge of skeleton topology is computed via Euclidean distance of two joint centroids aggregating all action classes, and each edge of action relation is calculated in the same manner by aggregating all joints dimension. The semantic relationships encoded by the LLM can enhance spatial-temporal modeling with graph convolution and improve the downstream classification performance.

\section{Future Research Opportunities}

Time series analysis with LLMs is an emerging and rapidly growing research area. Despite significant advances that have been made in the area, there are still many challenges, which open up a number of research opportunities. In this section, we highlight important research directions:

\begin{itemize}

    \item \textbf{Tokenization \& Prompt Design}. Tokenization plays a foundational role in capturing temporal dynamics of the input time series data. Existing techniques either rely on a single timestamp or patches of time steps to perform tokenization which could be insufficient to encode the time series for either general or specific applications. Therefore, it is important to develop novel tokenization methods that can better capture the temporal dynamics and facilitate underlying applications. For instance, \cite{rasul2023lagllama} employed lag features where the lags are derived from a set of appropriate lag indices for quarterly, monthly, weekly, daily, hourly, and second-level frequencies that correspond to the frequencies in the corpus of time series data. Based on appropriate tokenization, it is equally important to investigate how to design better prompts to further improve the model performance. For instance, we may develop a prompt learning architecture based on ~\cite {wang2022learning} to learn better prompts for specific tasks.

    \item \textbf{Interpretability}. Existing methods for LLMs based time series analysis aim to develop better tokenization, prompt design, fine-tune strategy, and integrate them to improve the model performance. However, these models are typically black-box, and therefore their output lacks explainability. In some applications, it is critical to explain the rationale of the model output to make it trustworthy. For this purpose, we may explore prototype-based methods and gradient-based methods to provide interpretations for the LLMs' output. We may also leverage knowledge distillation to train an explainable student model~\cite{mai2023enhanced} to enhance the interpretability of LLMs.
    
    \item \textbf{Multi-modality}. Time series data could be associated with data from other sources. For instance, in healthcare, we may not only collect the continuously monitored heart rate and blood pressure (time series) and medical records (texts and tabular data) but also collect X-rays (images). In this case, it is important to investigate how to incorporate multi-modality input via LLMs, how to align different modalities of input in the embedding space, and how to interpret the output accordingly. 
    
    \item \textbf{Domain Generalization}. One of the key challenges for LLM-based time series analysis is domain generalization which aims to generalize the model learned from one or more source domains to unseen target domains. Therefore, it is essential to tackle the distribution shift or domain shift problem by leveraging appropriate time series augmentation techniques,  learning temporal features that are invariant across domains (\textit{i.e.}, shared temporal dynamics or structures that are common to all domains), or meta-learning which aims to rapidly adapt to new time series tasks with limited examples from the target domain.

    \item \textbf{Scaling Laws of Time Series LLMs}. One critical research direction over LLMs is to understand their scaling laws, which aim to learn the patterns that depict how the increment of LLMs' size (\textit{e.g.}, in terms of the number of parameters) may impact their performance. Based on time series data, it is also crucial to verify whether the existing scaling laws are still valid based on either zero-shot learning, prompt learning, fine-tuning LLMs, or integration which will be tailored to specific time series tasks and applications.

    \item \textbf{Time Series LLMs as Agents}. LLM-based time series analysis can capture the temporal dynamics of the input time series and therefore can be used to assist in decision-making processes. By analyzing large volumes of time series data and their associated actions or rewards, LLMs can predict the outcomes based on historical data and summarize potential options based on the current status. As agents, time series LLMs can be adapted based on user preference, history, or context, to provide more personalized prediction and decisions. They can also serve as intermediaries or facilitators to seamlessly integrate with various systems and data sources to gather pertinent information, initiate actions, and deliver more extensive services.

    \item \textbf{Bias and Safety}. LLMs are trained on large-scale datasets collected from the internet and other sources which could inevitably involve bias in the training data. Because of this, LLMs may not only replicate but also amplify biases. To mitigate this issue, we should consider including a diverse range of data in the training set to reduce potential biases. We may also develop algorithms to detect, assess, and correct potential biases in the LLMs' output. Meanwhile, it is critical for time series LLMs to provide accurate and reliable output, especially in mission-critical systems such as healthcare and power systems. We should conduct rigorous tests over a wide range of potential scenarios to ensure the reliability and safety of LLMs' outputs. Furthermore, we should continuously track the LLMs' performance in the practical applications and incorporate user feedback to enhance safety and reliability.

\end{itemize}

\section{Conclusion}
In this survey, we provide a detailed overview of existing time series LLMs. We categorize and summarize the existing methods based on the proposed taxonomy of methodology. We also thoroughly discuss the applications of time series LLMs and highlight future research opportunities to advance time series analysis.


 
\bibliographystyle{named}
{{\footnotesize\bibliography{ijcai24}}}

\end{document}